\documentclass[11pt]{article}
\usepackage[final]{acl}

\usepackage{times}
\usepackage{latexsym}
\usepackage[T1]{fontenc}
\usepackage[utf8]{inputenc}
\usepackage{microtype}
\usepackage{inconsolata}
\usepackage{graphicx}

\usepackage{amsmath}
\usepackage{multicol}
\usepackage{multirow}
\usepackage{pifont}
\usepackage[most]{tcolorbox}
\usepackage{colortbl}
\usepackage{setspace}
\usepackage{booktabs}
\usepackage{float}
\usepackage{makecell}
\usepackage{cleveref}

\newcommand{\findingz}[2]{
    \begin{tcolorbox}[
        colback=blue!5,
        colframe=blue!60!black,
        arc=2pt,
        boxsep=2pt,
        left=5pt, right=5pt,
        top=2pt, bottom=2pt,
        boxrule=0.6pt,
        drop shadow={opacity=0.25},
        enhanced jigsaw
    ]
    \textbf{\textit{Finding #1:}} #2
    \end{tcolorbox}
}

\title{CAPruner: Conceptual-Adjacent Scene Graph Pruner for Enhancing\\3D Spatial Reasoning of Large Language Models}

\author{
    \textbf{Shengli Zhou\textsuperscript{1}},
    \textbf{Xiangchen Wang\textsuperscript{1}},
    \textbf{Guanhua Chen\textsuperscript{1*}},
    \textbf{Feng Zheng\textsuperscript{1,2}\thanks{Co-corresponding authors.}}
    \\
    \textsuperscript{1}Southern University of Science and Technology
    \textsuperscript{2}SpatialTemporal AI
    \\
    \small\texttt{\{zhousl2022, wangxc2019\}@mail.sustech.edu.cn, chengh3@sustech.edu.cn, f.zheng@ieee.org}
}

\begin{document}
\maketitle

\begin{abstract}
Large language models (LLMs) have recently been applied to 3D vision-language (3D-VL) tasks, which require spatial reasoning to identify target objects relative to anchors. Scene graphs are commonly employed to represent such relations, but reasoning over complete graphs incurs high token costs and computational inefficiencies, motivating the need for pruning.
Existing pruning methods primarily rely on spatial proximity and often remove task-relevant relations, thereby undermining reliable spatial reasoning.
To address these limitations, we derive a key requirement for scene graph pruning: preserving spatial relations that are most pertinent to the specific 3D-VL task. Guided by this insight, we propose the \textbf{C}onceptual-\textbf{A}djacent Scene Graph \textbf{Pruner} (\textbf{CAPruner}). CAPruner integrates fuzzy semantic relevance with spatial proximity to estimate the importance of relations, enabling the selection of critical relations in a task-specific context. Moreover, to avoid costly relation-level annotations, CAPruner is trained by supervising the aggregated scores of each node's incident edges. Extensive experiments demonstrate that CAPruner effectively preserves relations essential for spatial reasoning, leading to substantial performance improvements of LLMs on 3D-VL tasks. 
Code is available at \url{https://github.com/fz-zsl/CAPruner}.
\end{abstract} 
\section{Introduction}

\begin{figure}[t]
    \centering
    \includegraphics[width=0.48\textwidth]{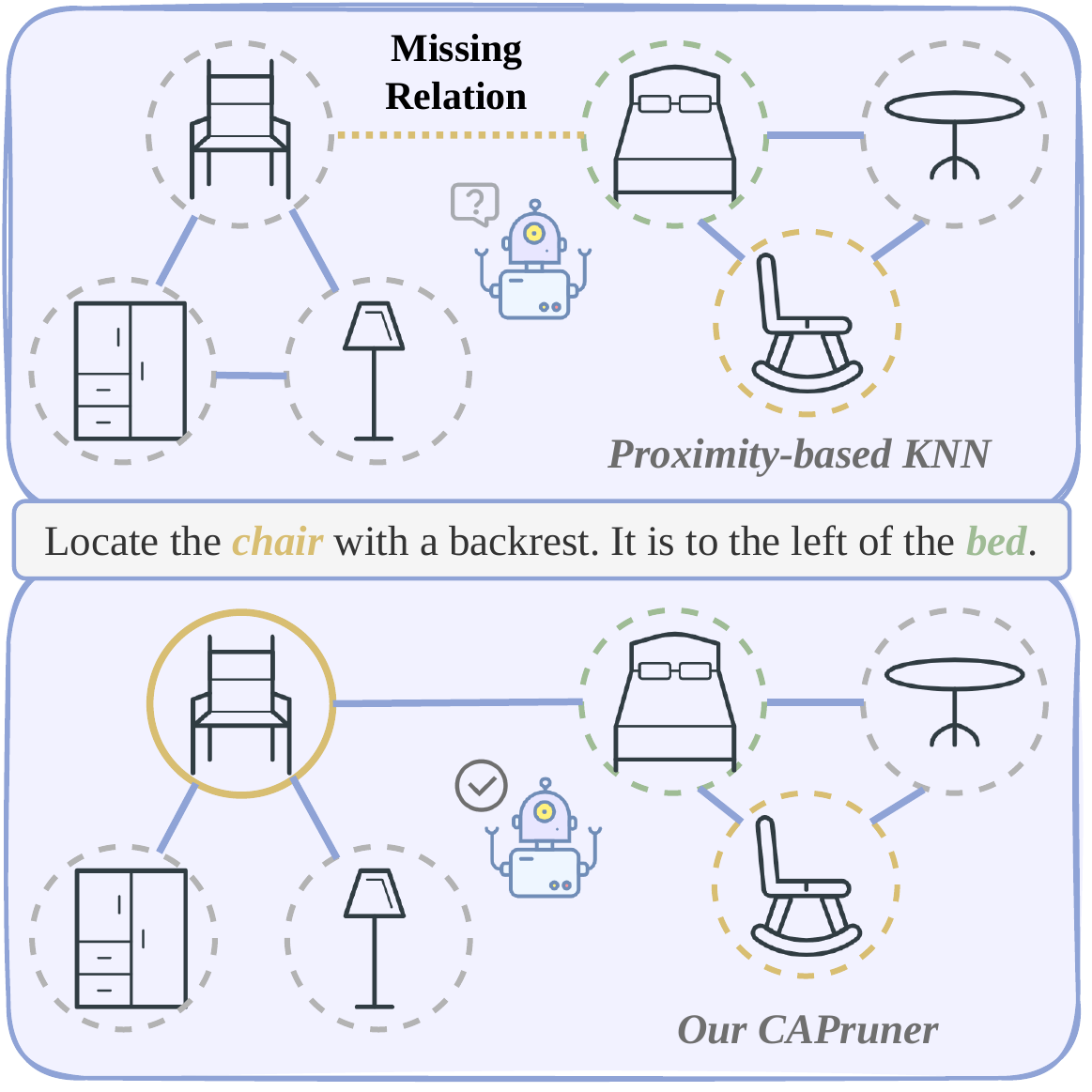}
    \caption{Comparison of scene-graph pruning strategies for LLM spatial reasoning. Previous proximity-based KNN keeps only nearest-neighbor relations and may discard task-critical long-range relations. CAPruner instead combines query-aware semantic cues with spatial proximity to preserve relations that are more useful for downstream reasoning.}
    \label{fig:teaser}
\end{figure}

With the development of large language models (LLMs) and their improved reasoning ability, using pretrained LLMs to assist spatial reasoning has become a new paradigm for solving 3D Vision-Language (3D-VL) tasks \cite{3dllm,leo,chat-scene,3dgraphllm}. These tasks require identifying a target object based on its relative position to anchor objects, making accurate perception of spatial relationships crucial for effective spatial reasoning. To better represent relative positions among objects, prior work introduced scene graphs (i.e., an abstract graph whose nodes are objects and whose edges represent relative position relations between objects) for LLMs to perceive the scene. However, as $n$ objects produce $\binom{n}{2}$ pairwise relations, sending all relation descriptions directly to an LLM causes a huge token count. This significantly reduces reasoning efficiency and hinders the extraction of useful information, making scaling impractical. For example, in the InteriorGS dataset \cite{interiorgs}, the average number of objects per scene exceeds 554, which would make the input token count reach the million level and exceed the input length limits of many LLMs. In real-world scenarios with even more objects, encoding all pairwise relations becomes even less feasible. To address this, 3DGraphLLM \cite{3dgraphllm} uses a proximity-based K-Nearest-Neighbors (KNN) pruning strategy and encodes only the relations between each object and its two nearest neighboring objects.

However, as shown in Fig. \ref{fig:teaser}, proximity is not strongly correlated with the necessity of a specific binary relation for solving a given task. Combined with the sparsity that KNN pruning produces in the scene graph, this approach cannot guarantee that the relations required to solve a problem are preserved after pruning. When a necessary relation (e.g., the relation between the bed and the chair in the figure) is pruned, the LLM cannot use the anchor object to find the target. Moreover, such an approach does not guarantee the connectivity of the remaining scene graph. Thus, proximity-based KNN cannot fully capture the layout of the entire scene, as the relations between different connected components are missing. As LLMs rely heavily on the retained relations, both shortcomings lead to errors in spatial reasoning.

These observations lead to an important question: \textbf{Under a limited budget, which relations in the scene graph should be kept?} In 3D-VL tasks, textual descriptions of relations between objects may include references to anchor objects using their category names and a spatial relation, e.g., ``locate the chair (target) next to (relation) the table (anchor)''. Hence, we claim that the importance of a relation can be measured by the attributes of both the incident objects and their positional correlation.

Based on this, we propose a lightweight \textbf{Conceptual-Adjacent Scene Graph Pruner (CAPruner)} to select object relations that are potentially useful for solving specific 3D-VL tasks. To reduce the risk of mistakenly pruning relations needed to answer the query, fuzzy matching is applied. For each object, we assign its weight by computing the semantic similarity between its name and words in the natural-language description of a specific task. For example, for the phrase ``red chair'', we give higher weight to all chairs in the scene without checking each chair's actual color. This reduces pruning mistakes that could deteriorate downstream LLM reasoning. For spatial relations between objects, the type of relation (e.g., left, right) can depend on the viewpoint and is hard to accurately determine by a compact pruning model. Hence, we weight edges by object distance following the Maxim of Relation \cite{logic_and_conv}. Combining these two factors, CAPruner computes the importance of each edge in the scene graph. The network takes the semantic similarity score of the two endpoint objects and their distance as input and outputs an edge weight for pruning.

\textbf{During training}, as current 3D-VL datasets only annotate the target object, and annotating all pairwise relations is prohibitively expensive, we supervise edge-weight learning using only the available target object labels. Concretely, we aggregate the weights of edges around each node and use the aggregated node score for supervision. This encourages edges near the target object to receive higher weights. \textbf{At pruning time}, we compute edge weights in the same way for each scene. For every node in the scene graph, we keep incident edges with the highest weights. Since the pruned scene graph only needs to preserve relations required by solving a specific task (i.e., not to fully describe the entire scene), our method avoids the trade-off between preserving global graph connectivity and retaining query-relevant relations.

To sum up, our main contributions are:
(1) We conduct qualitative experiments that reveal LLMs perform poorly on spatial relations that are not explicitly mentioned. From this, we summarize the scene-graph requirements when using LLMs for spatial reasoning.
(2) We propose CAPruner, a lightweight scene-graph pruning model that uses fuzzy matching on semantics and spatial proximity, plus aggregated node supervision for edge weights. It preserves limited pairwise relations while keeping relations needed to answer a given 3D-VL task.
(3) We validate the pruning rationale through extensive experiments. Both quantitative and qualitative results show that our method helps downstream LLMs perform spatial reasoning more accurately.
\section{Related Work}

\subsection{3D LLMs for 3D-VL Tasks}

In spatial reasoning tasks, models must accurately perceive the spatial relationships between objects to generate correct answers. Given the limited availability of 3D scene-text paired data, prior research has leveraged the perception and reasoning capabilities of large language models (LLMs) to enhance spatial reasoning. One such approach, 3D-LLM \cite{3dllm}, encodes the entire 3D scene as a holistic feature. While this method preserves the general layout of the scene, it sacrifices fine-grained details and mixes features of all objects, making it difficult for the model to identify individual objects and hindering object-level spatial reasoning. To address this, LEO \cite{leo} and Chat-Scene \cite{chat-scene} segment the scene into distinct objects, encoding the features of each object as input tokens. Although promising, these methods struggle to effectively extract spatial relations between objects, as the absolute positions are prematurely fused with geometric features.

To overcome these limitations, 3DGraphLLM \cite{3dgraphllm} introduces additional input tokens to explicitly represent spatial relations between objects. However, as the number of relations grows quadratically with object count (often exceeding the input limits of LLMs), 3DGraphLLM adopts a proximity-based KNN strategy, encoding only the relations between each object and its two nearest neighbors. Despite this, the approach remains prone to errors, as it neglects task-specific context and proximity does not always align with task-relevant importance.
In contrast, CAPruner considers both semantic relevance and spatial layout when pruning scene graphs, providing LLM with key relations for solving specific 3D-VL tasks.

\subsection{Scene Graphs}

Scene graphs are structured representations that capture objects and the semantic relations between them. Originally developed in the 2D vision-language domain, scene graphs have been effective for tasks such as image retrieval, referring expression comprehension, and captioning. Extending these concepts to 3D provides a promising way to incorporate structured, relational knowledge in 3D vision-language reasoning.

Scene graphs have also been widely adopted in the 3D domain to address robotics-oriented challenges, including motion planning \cite{sg20,sg53}, object localization for navigation \cite{sg17,sg20,sg34,sg53}, and robotic manipulation \cite{sg20}, as well as 3D scene construction \cite{sg16,sg59}. Aligned with the fast-growing trend of integrating scene graphs for enhancing spatial reasoning in various 3D-VL tasks, several previous methods have leveraged scene graphs to represent spatial relations between objects in the scene. OVSG \cite{ovsg} frames the 3D visual grounding problem as a subgraph retrieval task; 3DGraphQA \cite{3dgraphqa} facilitates 3D visual question-answering by introducing a bilinear graph neural network to realize feature fusion between scene graphs and question graphs; FFL-3DOG \cite{ffl3dog} constructs scene graphs based on a textual and visual information and aligns them to obtain the target.
Despite these methods having achieved promising results in their respective 3D-VL tasks, they fail to design task context-specific scene graphs tailored to the demands of LLMs (as discussed in Sec. \ref{sec:findings}), and thus are prone to omitting critical spatial relations that are essential for robust LLM-driven spatial reasoning.

In contrast, CAPruner provides task context-based scene graph pruning. It retains critical spatial relations tailored to LLM reasoning requirements while discarding redundant structural information, thereby enhancing the efficiency and accuracy of LLM-driven spatial reasoning in 3D-VL tasks.
\section{Findings} \label{sec:findings}

In this section, we explore the role of scene graphs in spatial reasoning tasks for LLMs. Specifically, we have addressed the following two core questions: (1) Why is the construction of scene graphs critical for subsequent reasoning tasks? (2) Why do existing scene graph pruning methods exhibit significant issues?

\begin{figure*}[t]
    \centering
    \includegraphics[width=0.95\textwidth]{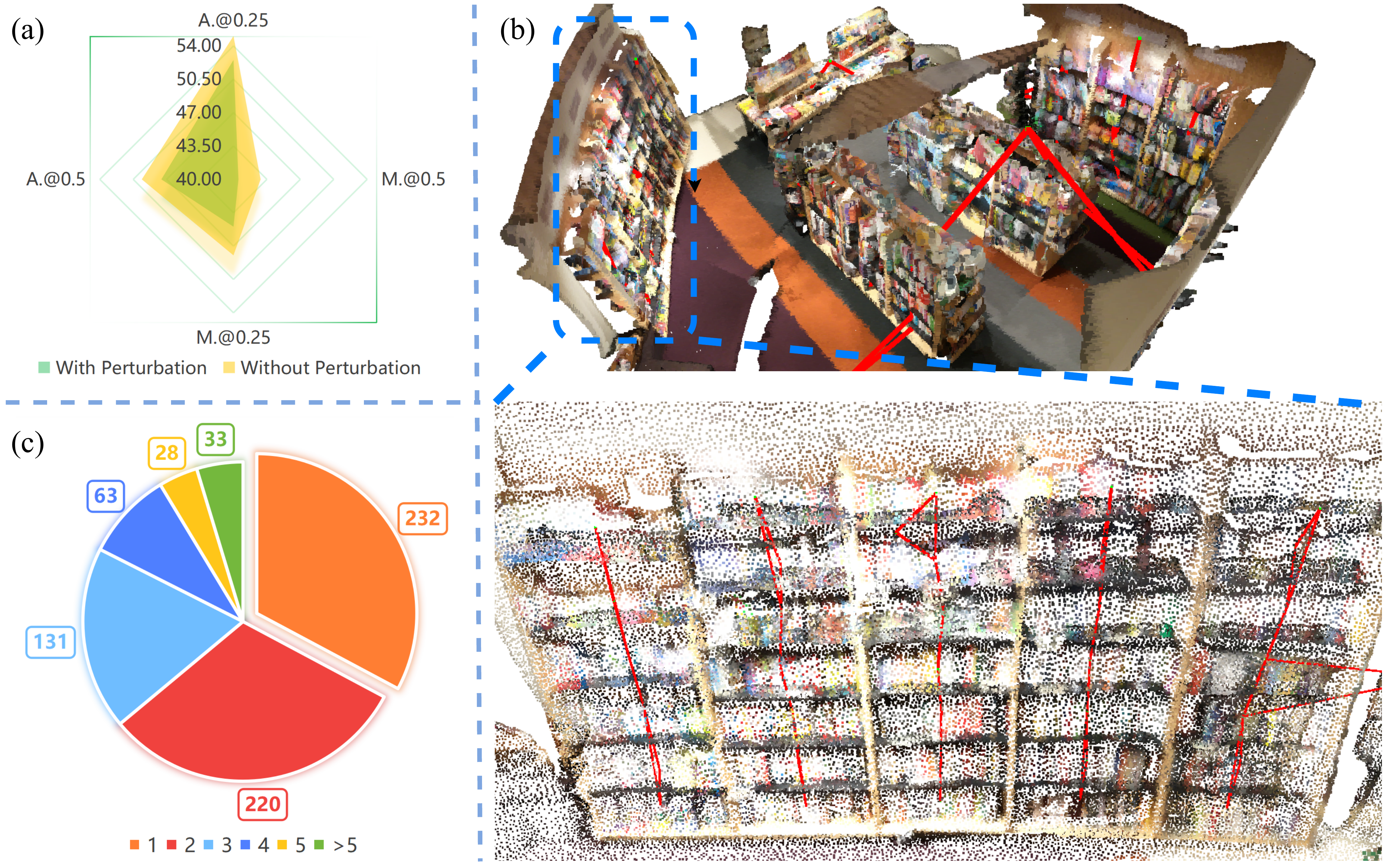}
    \caption{Our findings. (a) Replacing task-critical relations in the scene graph with irrelevant ones consistently degrades downstream accuracy, showing that LLM spatial reasoning relies heavily on the retained relations. (b) Proximity-based KNN overlooks the importance of the context of individual tasks, causing redundancy in local regions (lower part) and global insufficiency (upper part). (c) Over $67\%$ of ScanNet scenes become disconnected after proximity-based KNN pruning, weakening the graph's ability to represent the global scene layout.}
    \label{fig:findings}
\end{figure*}

\subsection{Importance of Scene Graph Pruning}

To demonstrate the importance of scene graph pruning for LLMs in spatial reasoning tasks, we compare the effects of different edge-selection strategies on downstream task performance. Specifically, we fine-tune LLMs with well-pruned scene graphs and perturbed scene graphs (where some critical spatial relations are removed manually). Then, we test the models on the validation split of the ScanRefer dataset \cite{scanrefer}, which has demanding spatial reasoning requirements, and compare the accuracy of the LLM's responses.

As shown in \cref{fig:findings} (a), model performance declines after perturbation, indicating that the absence of key relations leads to insufficient perception of the scene, resulting in errors in spatial reasoning. This underscores the model's reliance on the relative positioning of objects encoded in the scene graph. Therefore, to better support spatial reasoning, the pruning model must ensure that spatial relations essential for addressing specific tasks are preserved in the pruned edges.

\findingz{1}{Removing key spatial relationships significantly impairs spatial reasoning.}

\subsection{Shortcomings of Current Pruners}

Existing methods, such as 3DGraphLLM \cite{3dgraphllm}, employ a proximity-based KNN pruning strategy, which preserves the spatial relations between each object and its two nearest neighbors in the scene. Such an approach defines the importance of relations solely based on the distance between objects. However, this approach has two key limitations:

\textbf{Semantic and Structural Gaps.}
As the importance of relations is determined entirely by the distance between objects, they neglect the importance of semantic information (e.g., object categories) in the context of specific 3D-VL tasks. For example, in Fig. \ref{fig:findings} (b), the proximity between bookshelves and books leads the proximity-based KNN pruning method to establish numerous connections between them, while overlooking relations with other objects. As a result, the LLM provides incorrect answers when tasked with locating ``the bookshelf with 7 shelves and 4 sections, located on the wall opposite the table that has books on it,'' as there is no relation between the bookshelf and the wall.
Furthermore, the proximity-based KNN pruning strategy does not ensure the connectivity of the pruned scene graph. As illustrated in Fig. \ref{fig:findings} (c), among the 707 scenes from the ScanNet data \cite{scannet}, only 232 (less than 33\%) retain connectivity after pruning. For other scenes, the scene graph fails to represent the relative positions of objects between connected components and, thus, is unable to represent the layout of the entire scene. This severely limits spatial reasoning and the model's ability to comprehend the scene.
\findingz{2}{Scene graph pruning should integrate semantic information and maintain structural integrity for the region of interest.}

\begin{figure*}[t]
    \centering
    \includegraphics[width=0.95\textwidth]{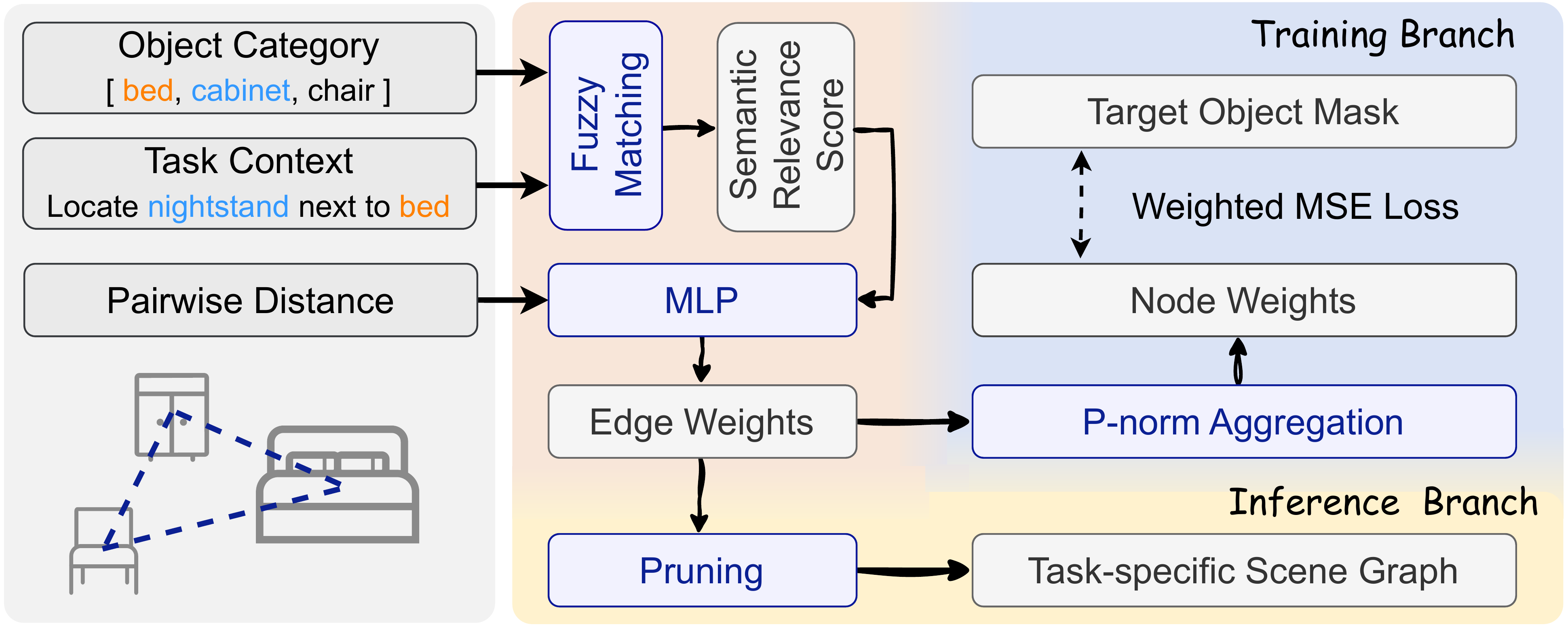}
    \caption{Overview of \textbf{CAPruner}. The framework first estimates object-query semantic relevance via fuzzy matching and combines it with geometric cues to predict edge weights. The training branch aggregates edge weights into node weights via p-norm aggregation for node-wise supervision through weighted MSE loss, while the inference branch prunes edges based on edge weights to generate a task-specific scene graph.}
    \label{fig:model}
\end{figure*}

\textbf{Rigid Pruning Policies.}
Previous methods \cite{vlsat,3dgraphllm} rely on fixed pruning strategies to select relations for representing the layout of objects within a scene. Such an approach limits their flexibility and ability to adapt to the specific context of a query. This rigidity hinders the model's ability to prioritize the most context-relevant information, forcing it to encode unnecessary relations. For example, when tasked with a 3D visual grounding query identifying ``the table to the north of the two bookshelves in the center of the room, which is a long, creamy brown rectangle'' in the scene shown in Fig. \ref{fig:findings} (b), the fixed pruning strategy fails to emphasize the relations crucial to the query (e.g., table-bookshelf). Instead, it retains redundant connections, such as those between sections of bookshelves, wasting budget on irrelevant relations. Consequently, the model inefficiently uses computational resources by processing irrelevant information that does not contribute to the specific task at hand.

\findingz{3}{Pruning strategy needs to be adaptive to task-specific context for prioritizing relevant information.}

Based on the analysis above, we can propose a basic paradigm for scene graph pruning algorithms: (1) To better extract scene structural information for LLM spatial reasoning, it is essential to retain the spatial relations between objects that are crucial for answering specific 3D-VL tasks in the pruned scene graph; (2) During scene graph construction, the pruning model should integrate the context of specific 3D-VL tasks, the semantic information of objects, and the spatial relations between objects to assess the importance of each edge.

\section{Method}

\subsection{Fuzzy Matching}

According to the above-mentioned paradigm, we propose \textbf{Conceptual-Adjacent Scene Graph Pruner (CAPruner)}.
CAPruner is designed to prioritize relations that are crucial in the context of specific 3D-VL tasks according to the semantic properties of objects and their spatial relations in the scene. While LLMs are highly sensitive to missing key relations, they are relatively robust to small amounts of redundant information. Thus, the goal of CAPruner is to minimize the risk of erroneous pruning under a limited budget of retained edges.

For individual objects, textual references typically include the object's name (e.g., ``table'', ``cup'') and its properties (e.g., ``rectangular'', ``red''). Due to current limitations in aligning 3D point cloud features with textual data \cite{fuzzycateg1,fuzzycateg2}, filtering referents according to the object's detailed properties can easily lead to incorrect pruning decisions.
For example, when attempting to match a ``red, round table next to three chairs,'' any error in determining the object's color, shape, or relative positioning can result in the actual referent being mistakenly deemed unimportant, triggering false rejection in pruning.
To address this, we match objects semantically based on their categories. For instance, when the query mentions a ``table,'' the scene's table and semantically similar objects (e.g., ``desk'') receive extra weight according to their semantic similarity.

Regarding the spatial relationships between objects, many relational expressions are perspective-dependent (e.g., ``front'', ``back'', ``left'', and ``right''). Such a property has made models struggle with understanding these relational categories \cite{fuzzyrel}. To prevent incorrect pruning in such cases, we assign higher weights to objects that are closer in proximity, in accordance with the Maxim of Relation theory \cite{logic_and_conv}.

\subsection{Model Architecture}

In this section, we introduce the architecture of the CAPruner model, as shown in Fig. \ref{fig:model}. The backbone of CAPruner first calculates semantic relevance scores, and then obtains the weight of each edge as a function of the distance and semantic relevance scores of the objects at its ends. The training branch aggregates edge weights to node weights for node-wise supervision, while the inference branch prunes according to edge weights.

\textbf{Semantic Relevance Calculation.}
The semantic relevance of an object is determined by the degree to which its category aligns with the context of the 3D-VL task description. Let $\mathcal T$ represent the set of tokens in the natural language description of the 3D-VL task, and $c_i$ denote the category of object $i$. The semantic relevance score $s_i$ for object $i$ is calculated as the maximum similarity between the object's category and the task tokens, i.e.,

\begin{equation}
   s_i=\max_{t\in\mathcal T}\left\{\text{Similarity}(c_i, t)\right\}
\end{equation}

\textbf{Edge Weight Calculation.}
The weight of an edge between two objects $i$ and $j$ is determined by their semantic relevance scores and spatial proximity within the scene. Let $P_i$ and $P_j$ denote the positions of objects $i$ and $j$ in the 3D scene. The edge weight $w_{ij}$ is defined as a function of the semantic relevance scores $s_i, s_j$ and the Euclidean distance $||P_i-P_j||_2$ between the objects. Formally, $w_{ij} = f(s_i, s_j, ||P_i - P_j||_2)$, where $f(\cdot)$ is a multi-layer perceptron that aggregates these features to compute the edge weight.

\textbf{Node-wise Supervision.}
Currently, many mainstream 3D-VL datasets \cite{scanrefer, scanqa, sqa3d} provide annotations for the target object with respect to the textual description. However, none of them provides the importance of edges between each pair of objects. Moreover, as the labeling effort of such annotations scales quadratically with the number of objects, it is impractical to manually label task-specific inter-object relations. Hence, CAPruner employs a node-wise supervision that aggregates edge weights to node weights, supervises node weights, and propagates the effect back to edge weights.
Specifically, the weight of each node is calculated using a graph neural network (GNN) approach that aggregates the weights of the edges incident to it. This aggregation is performed using the p-norm method $v_i = \text{sigmoid}\left(\sum_j w_{ij}^p\right)^{1/p}$, where $v_i$ is the node weight of the $i$-th object, and $p$ is a hyperparameter that controls the p-norm aggregation.

\begin{table*}[!ht]
    \centering
    \scalebox{0.85}{
    \begin{tabular}{l|cccc|c|c}
    \toprule
        \multirow{2}{*}{\textbf{Model}} & \multicolumn{4}{c|}{\textbf{ScanRefer}} & \textbf{ScanQA} & \textbf{SQA3D} \\ 
        ~ & A.@0.25 & A.@0.5 & M.@0.25 & M.@0.5 & BLEU-4 & EM@1 \\ \midrule
        FFL-3DOG~\cite{ffl3dog} & 41.3 & 34.0 & 35.2 & 25.7 & -- & -- \\ 
        SeeGround~\cite{seeground} & 44.1  & 39.4  & 34.0  & 30.0  & -- & -- \\ 
        CSVG~\cite{csvg} & 49.6  & 39.8  & 38.4  & 27.3  & -- & -- \\ 
        AugRefer~\cite{augrefer} & 55.7  & 44.0  & 50.0  & 39.1  & -- & -- \\ 
        MA2TransVG~\cite{ma2transvg} & 57.9  & 45.7  & 53.8  & 41.4  & -- & -- \\ 
        3D-VisTA~\cite{3dvista} & 50.6  & 45.8  & 43.7  & 39.1  & 13.1  & 48.5  \\ 
        TSP3D~\cite{tsp3d} & 56.5  & 46.7  & -- & -- & -- & -- \\ 
        Scene-LLM~\cite{scenellm} & -- & -- & -- & -- & 12.0  & 54.2  \\ 
        Chat-Scene-7B~\cite{chat-scene} & 55.5  & 50.2  & -- & -- & 14.3  & 54.6  \\ 
        PQ3D~\cite{pq3d} & -- & 51.2  & -- & 46.2  & -- & 47.1 \\ 
        QuatRoPE~\cite{quatrope} & 58.2 & 52.5 & 54.3 & 49.2 & -- & 55.2 \\  \midrule
        3DGraphLLM-1B~\cite{3dgraphllm} & 52.5  & 47.5  & 45.0  & 40.5  & 12.2  & 52.6  \\ 
        \rowcolor{gray!20} \textbf{CAPruner + Llama-3.2-1B (Ours)} & 55.0  & 49.6  & 48.0  & 42.8  & 13.0  & 52.8  \\ 
        3DGraphLLM-8B~\cite{3dgraphllm} & 60.2  & 54.6  & \textit{54.7}  & \textit{49.4}  & 12.5  & 55.2  \\ 
        \rowcolor{gray!20} \textbf{CAPruner + Llama-3-8B (Ours)} & 61.7  & 56.0  & 55.3  & 49.9  & 13.2  & 56.3 \\ \bottomrule
    \end{tabular}
    }
    \caption{Comparison on ScanRefer, ScanQA, and SQA3D. With the same backbone LLM, CAPruner consistently outperforms base methods and achieves the strongest or highly competitive results. A. and M. denote accuracy on the overall and ``multi'' splits, respectively. Scores for 3DGraphLLM-1B and in italic are evaluated on our machine.}
    \label{tab:exp_main}
\end{table*}

\textbf{Weighted MSE loss.}
Since the number of target objects in the 3D-VL task is typically much smaller than the number of non-target objects, we balance the contributions of target and non-target objects using a weighted mean squared error (WMSE) loss. For the target object set $\mathcal{O}$ and the non-target object set $\widetilde{\mathcal{O}}$, the loss function $\mathcal L$ is defined as:

\begin{equation} \label{eq:loss}
   \mathcal L=\frac{1}{|\mathcal{O}|}\sum_{i\in\mathcal{O}}(v_i-1)^2+\frac{1}{|\widetilde{\mathcal{O}}|}\sum_{i\in\widetilde{\mathcal{O}}} v_i^2
\end{equation}

\noindent As the sigmoid function limits $v_i$ in the range of $(0,1)$, the first term encourages nodes corresponding to target objects to have higher weights, and the second term encourages nodes corresponding to non-target objects to have lower weights. 

These weights are then propagated back through the p-norm aggregation operation on the scene graph, encouraging edges incident to target objects to have higher edge weights, while lowering the edge weights of edges incident to non-target objects.
Such a supervision approach helps the model to strike a better balance between semantic relevance and proximity.

By the end of training, the scene graph can be flexibly pruned according to edge weights with respect to the task context, preserving the most relevant relationships for the task at hand while discarding redundant or irrelevant connections.
\section{Experiments}

\subsection{Experimental Settings}

~~\textbf{Implementation Details.}
For semantic relevance calculation, we classify objects into general categories defined in the NYUv2 dataset \cite{nyuv2}. The object receives a semantic similarity score of 1 only when there exists an object of the same category in the textual description. When computing edge weights using the semantic relevance score and objects' pairwise distances, we utilize a 3-layer MLP with 1219 parameters, which yields high computational efficiency.
To enhance the robustness of our model while making a fair comparison with the previous proximity-based KNN model, 3DGraphLLM \cite{3dgraphllm}, we preserve two incident edges with the highest weights for each node in the scene graph.
In the experiments, Llama-3.2-1B is used for 1B models, and Llama-3-8B is used for 8B models \cite{llama3}. When parsing the pruned scene graph into LLM for inference, we arrange a series of tokens in a sequence. The sequence pattern follows the same setting in 3DGraphLLM, encoding the features of texts, individual objects, and selected relations.

\textbf{Training Approach.}
During training, we first train the CAPruner model for pruning the scene graph. The model is trained for $50$ epochs with a batch size of $16$, a learning rate of $10^{-3}$, and $p=3$ for p-norm aggregation.
Then, we fine-tune the LLM using LoRA with $r=16$ for 3 epochs with a batch size of 8 and a learning rate of $2\times 10^{-5}$.

\begin{table*}[!ht]
    \centering
    \scalebox{0.87}{
    \begin{tabular}{c|cccc|cc|ccc}
    \toprule
        \multirow{2}{*}{\textbf{Pruning Method}} & \multicolumn{4}{c|}{\textbf{ScanRefer}} & \multicolumn{2}{c|}{\textbf{ScanQA}} & \multicolumn{3}{c}{\textbf{SQA3D}} \\ 
        ~ & A.@0.25 & A.@0.5 & M.@0.25 & M.@0.5 & B.-3 & B.-4 & EM@1 & ROUGE & CIDEr \\ \midrule
        Proximity-based KNN & 52.5  & 47.5  & 45.0  & 40.5  & 17.9  & 12.2  & 52.6  & 53.8  & 138.3  \\ 
        CAPruner (MST) & 54.4  & 49.0  & 47.1  & 42.0  & 18.1  & 11.7  & 52.4  & 53.7  & 139.0  \\ 
        \rowcolor{gray!20} Gain & 1.9  & 1.5  & 2.1  & 1.5  & 0.2  & -0.5  & -0.2  & -0.1  & 0.7  \\ 
        CAPruner (KNN) & 55.0  & 49.6  & 48.0  & 42.8  & 18.5  & 13.0  & 52.8  & 54.1  & 139.1  \\ 
        \rowcolor{gray!20} Gain & 2.5  & 2.1  & 3.0  & 2.3  & 0.6  & 0.8  & 0.2  & 0.4  & 0.8 \\ \bottomrule
    \end{tabular}
    }
    \caption{Comparison of pruning policies after learning CAPruner edge weights. Applying KNN to the learned scores performs best. A. and M. denote accuracy on the overall and ``multi'' splits, respectively; B.-3 and B.-4 denote BLEU-3 and BLEU-4. All models are based on Llama-3.2-1B.}
    \label{tab:exp_abl_pru}
\end{table*}

\subsection{Comparative Experiment}

In this experiment, we aim to verify the effectiveness of CAPruner by comparing its performance against previous models. The experiments are conducted on the ScanRefer \cite{scanrefer} dataset for 3D visual grounding (3D VG), the ScanQA \cite{scanqa} dataset for 3D visual question-answering (3D VQA), and the SQA3D \cite{sqa3d} dataset for situated 3D VQA. 
We report localization accuracy on ScanRefer, BLEU-4 on ScanQA, and EM@1 on SQA3D.

The results in Tab. \ref{tab:exp_main} demonstrate that models using scene graphs pruned by CAPruner have achieved large gains throughout all metrics compared to models using proximity-based KNN pruning (i.e., 3DGraphLLM) when using the same LLM, especially on datasets with higher spatial reasoning demands like ScanRefer.
Our model has also achieved the highest scores for most metrics, showing the superiority of our method in spatial reasoning and comprehensive 3D-VL task-solving ability.
Additionally, CAPruner's inference time for each 3D-VL sample is 0.75 ms, which is negligible compared to the inference time of LLM (e.g., 1731 ms per sample for the 8B model).

\begin{figure}[t]
    \centering
    \includegraphics[width=0.48\textwidth]{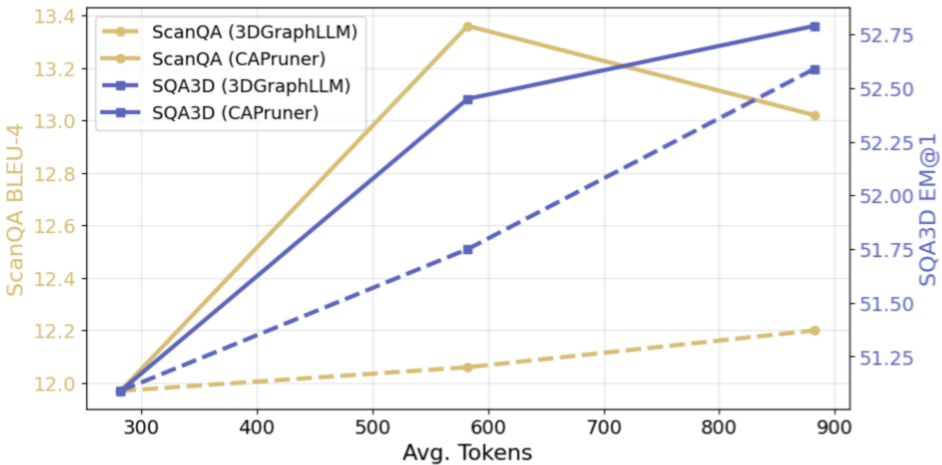}
    \caption{Comparison between CAPruner (1B) and  3DGraphLLM (proximity-based KNN, 1B)}
    \label{fig:save_token}
\end{figure}

Meanwhile, as shown in Fig. \ref{fig:save_token}, CAPruner not only outperforms proximity-based KNN across all token budgets but also achieves higher token efficiency for 3D-VL tasks. For example, CAPruner, with an average of 582 tokens, performs better than the baseline model, which uses an average of 882 tokens, saving 34\% in token usage.

\begin{table}[tp]
    \centering
    \scalebox{0.9}{
    \begin{tabular}{ccc}
    \toprule
        Proximity & Semantic Similarity & Accuracy \\ \midrule
        \ding{56} & None & 7.73  \\ 
        \ding{56} & Bert \cite{bert} & 8.21  \\ 
        \ding{56} & Strict Matching & 17.71  \\ 
        \ding{56} & Fuzzy Matching & 24.44  \\ \midrule
        \ding{52} & None & 7.73  \\ 
        \ding{52} & Bert \cite{bert} & 8.57  \\ 
        \ding{52} & Strict Matching & 24.38  \\ 
        \ding{52} & Fuzzy Matching & 24.57 \\ \bottomrule
    \end{tabular}
    }
    \caption{Ablation on proximity cues and semantic matching strategies. Strict matching helps when object categories are explicitly mentioned, but fuzzy matching is consistently stronger; combining fuzzy matching with proximity yields the best accuracy.}
    \label{tab:exp_abl_factor}
\end{table}

\subsection{Comparison on Pruning Approaches}

When using the scene graph to represent the layout of the entire scene, the connectivity after pruning is crucial to maintaining structural integrity, but it also hinders the model from focusing on more important relations. As CAPruner is designed to represent key relations in a specific 3D-VL task context, it only needs to focus on the region of interest, lowering the requirement on structural integrity.

In this experiment, after training the CAPruner model, we use KNN and MST (first select the edges on the maximum-weight spanning tree of the scene graph, then each node selects one more remaining incident edge with the largest weight) strategies for pruning. Finally, we fine-tune the downstream LLM and compare the performances.

The results in Tab. \ref{tab:exp_abl_pru} demonstrate that the KNN strategy performs better, verifying our advantage of forgoing the requirement of structural integrity by characterizing local regional features.

\subsection{Ablation Studies}

We perform ablation studies on model settings and measure the accuracy of CAPruner as the percentage of samples where it can correctly predict $|\mathcal O|$ (i.e., number of targets) objects with top node weights without involving LLM. Results for the effects of proximity, the semantic similarity computation method, and the choice of $p$ in p-norm aggregation are presented in Tab. \ref{tab:exp_abl_factor} and Tab. \ref{tab:ablp}.

The results lead to three conclusions: (1) Fuzzy matching outperforms other strategies, including Bert embeddings, whose discriminability is relatively low. (2) Proximity helps identify important edges and target objects, but its effect is much smaller than semantics, which also aligns with the Maxim of Relation that ``in reference, semantics are given priority, while proximal objects are considered when conditions are equal''. (3) Accuracy grows as $p$ grows for $p\leq 3$ and remains stable afterward, indicating that edges with large weights play a more important role than the sum of the edge weights in predicting target objects.

\begin{table}[tp]
    \centering
    \small
    \begin{tabular}{c|cccccc}
    \toprule
        $p$ & 1 & 2 & 3 & 4 & 5 & 6 \\ \midrule
        Acc. & 7.57 & 19.22 & 24.57 & 24.51 & 24.57 & 24.36 \\ \bottomrule
    \end{tabular}
    \caption{Ablation on $p$ used in p-norm aggregation.}
    \label{tab:ablp}
\end{table}
\section{Conclusion}

We presented CAPruner, a lightweight scene graph pruning model for LLM-based spatial reasoning in 3D-VL tasks. By combining task-specific semantic cues with spatial proximity and training with aggregated node-level supervision, CAPruner preserves relations that are most useful for downstream reasoning without requiring relation-level annotation. Experiments show that CAPruner consistently outperforms proximity-based pruning with negligible overhead, highlighting task-specific scene graph pruning as an effective and scalable strategy.
\section*{Limitations}

Though edges in scene graphs can correspond to most expressions in 3D-VL task descriptions, they cannot represent complex relations (e.g., the fourth chair counting from the left in a row of chairs behind the fifth row of desks in the classroom). Therefore, a more versatile and generalizable data structure should be considered to better represent complex relations in real-world scenarios.
\section*{Acknowledgements}

This work was supported by the National Key Research and Development Program of China under Grant 2024YFE0203100.

\bibliography{custom}

@String(CVPR= {IEEE Conf. Comput. Vis. Pattern Recog.})

@String(ICCV= {Int. Conf. Comput. Vis.})

@String(ECCV= {Eur. Conf. Comput. Vis.})

@String(BMVC= {Brit. Mach. Vis. Conf.})

@String(ICLR = {Int. Conf. Learn. Represent.})

@String(CVPR  = {CVPR})

@String(ICCV  = {ICCV})

@String(ECCV  = {ECCV})

@String(BMVC  =	{BMVC})

@String(ICLR  = {ICLR})

@article{chat-scene,
  title={Chat-scene: Bridging 3d scene and large language models with object identifiers},
  author={Huang, Haifeng and Chen, Yilun and Wang, Zehan and Huang, Rongjie and Xu, Runsen and Wang, Tai and Liu, Luping and Cheng, Xize and Zhao, Yang and Pang, Jiangmiao and others},
  journal={Proceedings of the Advances in Neural Information Processing Systems, Vancouver, BC, Canada},
  year={2024}
}

@misc{3dgraphllm,
      title={3DGraphLLM: Combining Semantic Graphs and Large Language Models for 3D Scene Understanding}, 
      author={Tatiana Zemskova and Dmitry Yudin},
      year={2024},
      eprint={2412.18450},
      archivePrefix={arXiv},
      primaryClass={cs.CV},
      url={https://arxiv.org/abs/2412.18450}, 
}

@inproceedings{scanrefer,
    title={Scanrefer: 3d object localization in rgb-d scans using natural language},
    author={Chen, Dave Zhenyu and Chang, Angel X and Nie{\ss}ner, Matthias},
    booktitle={Computer Vision--ECCV 2020: 16th European Conference, Glasgow, UK, August 23--28, 2020, Proceedings, Part XX 16},
    pages={202--221},
    year={2020},
    organization={Springer}
}

@InProceedings{multi3dref,
    author    = {Zhang, Yiming and Gong, ZeMing and Chang, Angel X.},
    title     = {Multi3DRefer: Grounding Text Description to Multiple 3D Objects},
    booktitle = {Proceedings of the IEEE/CVF International Conference on Computer Vision (ICCV)},
    month     = {October},
    year      = {2023},
    pages     = {15225-15236}
}

@inproceedings{sqa3d,
  title={SQA3D: Situated Question Answering in 3D Scenes},
  author={Ma, Xiaojian and Yong, Silong and Zheng, Zilong and Li, Qing and Liang, Yitao and Zhu, Song-Chun and Huang, Siyuan},
  booktitle={International Conference on Learning Representations},
  year={2023},
  url={https://openreview.net/forum?id=IDJx97BC38}
}

@inproceedings{scanqa,
  title={ScanQA: 3D Question Answering for Spatial Scene Understanding},
  author={Azuma, Daichi and Miyanishi, Taiki and Kurita, Shuhei and Kawanabe, Motoaki},
  booktitle={Proceedings of the IEEE/CVF Conference on Computer Vision and Pattern Recognition (CVPR)},
  year={2022}
}

@inproceedings{bert,
    title = "{BERT}: Pre-training of Deep Bidirectional Transformers for Language Understanding",
    author = "Devlin, Jacob  and
      Chang, Ming-Wei  and
      Lee, Kenton  and
      Toutanova, Kristina",
    editor = "Burstein, Jill  and
      Doran, Christy  and
      Solorio, Thamar",
    booktitle = "Proceedings of the 2019 Conference of the North {A}merican Chapter of the Association for Computational Linguistics: Human Language Technologies, Volume 1 (Long and Short Papers)",
    month = jun,
    year = "2019",
    address = "Minneapolis, Minnesota",
    publisher = "Association for Computational Linguistics",
    url = "https://aclanthology.org/N19-1423/",
    doi = "10.18653/v1/N19-1423",
    pages = "4171--4186"
}

@InProceedings{3dvista,
    author    = {Zhu, Ziyu and Ma, Xiaojian and Chen, Yixin and Deng, Zhidong and Huang, Siyuan and Li, Qing},
    title     = {3D-VisTA: Pre-trained Transformer for 3D Vision and Text Alignment},
    booktitle = {Proceedings of the IEEE/CVF International Conference on Computer Vision (ICCV)},
    month     = {October},
    year      = {2023},
    pages     = {2911-2921}
}

@misc{llama3,
      title={The Llama 3 Herd of Models}, 
      author={Aaron Grattafiori and Abhimanyu Dubey and Abhinav Jauhri and others},
      year={2024},
      eprint={2407.21783},
      archivePrefix={arXiv},
      primaryClass={cs.AI},
      url={https://arxiv.org/abs/2407.21783}, 
}

@article{dinov2,
    title={{DINO}v2: Learning Robust Visual Features without Supervision},
    author={Maxime Oquab and Timoth{\'e}e Darcet and Th{\'e}o Moutakanni and others},
    journal={Transactions on Machine Learning Research},
    issn={2835-8856},
    year={2024},
    url={https://openreview.net/forum?id=a68SUt6zFt},
    note={Featured Certification}
}

@article{mask3d,
    title   = {{Mask3D: Mask Transformer for 3D Semantic Instance Segmentation}},
    author  = {Schult, Jonas and Engelmann, Francis and Hermans, Alexander and Litany, Or and Tang, Siyu and Leibe, Bastian},
    journal = {{International Conference on Robotics and Automation (ICRA)}},
    year    = {2023}
}

@InProceedings{ma2transvg,
    author       = {Xu, Can and Han, Yuehui and Xu, Rui and Hui, Le and Xie, Jin and Yang, Jian},
    title        = {Multi Attributes Interactions Matters for 3D Visual Grounding},
    booktitle    = {CVPR},
    year         = {2024},
}

@article{augrefer,
    author       = {Xinyi Wang and Na Zhao and Zhiyuan Han and Dan Guo and Xun Yang},
    title        = {AugRefer: Advancing 3D Visual Grounding via Cross-Modal Augmentation and Spatial Relation-based Referring},
    journal      = {CoRR},
    volume       = {abs/2501.09428},
    year         = {2025},
    url          = {https://doi.org/10.48550/arXiv.2501.09428},
    doi          = {10.48550/ARXIV.2501.09428},
    eprinttype   = {arXiv},
    eprint       = {2501.09428},
    bibsource    = {dblp computer science bibliography, https://dblp.org}
}

@InProceedings{tsp3d,
    author    = {Guo, Wenxuan and Xu, Xiuwei and Wang, Ziwei and Feng, Jianjiang and Zhou, Jie and Lu, Jiwen},
    title     = {Text-guided Sparse Voxel Pruning for Efficient 3D Visual Grounding},
    booktitle = {Proceedings of the Computer Vision and Pattern Recognition Conference (CVPR)},
    month     = {June},
    year      = {2025},
    pages     = {3666-3675}
}

@InProceedings{scenellm,
    author    = {Fu, Rao and Liu, Jingyu and Chen, Xilun and Nie, Yixin and Xiong, Wenhan},
    title     = {Scene-LLM: Extending Language Model for 3D Visual Reasoning},
    booktitle = {Proceedings of the Winter Conference on Applications of Computer Vision (WACV)},
    month     = {February},
    year      = {2025},
    pages     = {2195-2206}
}

@inproceedings{leo,
  title={An Embodied Generalist Agent in 3D World},
  author={Huang, Jiangyong and Yong, Silong and Ma, Xiaojian and Linghu, Xiongkun and Li, Puhao and Wang, Yan and Li, Qing and Zhu, Song-Chun and Jia, Baoxiong and Huang, Siyuan},
  booktitle={Proceedings of the International Conference on Machine Learning (ICML)},
  year={2024}
}

@article{3dllm,
 author = {Hong, Yining and Zhen, Haoyu and Chen, Peihao and Zheng, Shuhong and Du, Yilun and Chen, Zhenfang and Gan, Chuang},
 title = {3D-LLM: Injecting the 3D World into Large Language Models},
 journal = {NeurIPS},
 year = {2023},
}

@article{logic_and_conv,
  title={Logic and conversation},
  author={H. Paul Grice},
  journal={Syntax and Semantics},
  year={1975},
  volume={3},
  pages={41-58},
  url={https://api.semanticscholar.org/CorpusID:222385009}
}

@misc{interiorgs,
  title        = {InteriorGS: A 3D Gaussian Splatting Dataset of Semantically Labeled Indoor Scenes},
  author       = {SpatialVerse Research Team, Manycore Tech Inc.},
  year         = {2025},
  howpublished = {\url{https://huggingface.co/datasets/spatialverse/InteriorGS}}
}

@InProceedings{pq3d,
    author="Zhu, Ziyu
    and Zhang, Zhuofan
    and Ma, Xiaojian
    and Niu, Xuesong
    and Chen, Yixin
    and Jia, Baoxiong
    and Deng, Zhidong
    and Huang, Siyuan
    and Li, Qing",
    editor="Leonardis, Ale{\v{s}}
    and Ricci, Elisa
    and Roth, Stefan
    and Russakovsky, Olga
    and Sattler, Torsten
    and Varol, G{\"u}l",
    title="Unifying 3D Vision-Language Understanding via Promptable Queries",
    booktitle="Computer Vision -- ECCV 2024",
    year="2025",
    publisher="Springer Nature Switzerland",
    address="Cham",
    pages="188--206",
    isbn="978-3-031-72784-9"
}

@inproceedings{csvg,
    author    = {Qihao Yuan and Kailai Li and Jiaming Zhang},
    title     = {Solving Zero-Shot 3D Visual Grounding as Constraint Satisfaction Problems},
    booktitle = {36th British Machine Vision Conference 2025, {BMVC} 2025, Sheffield, UK, November 24-27, 2025},
    publisher = {BMVA},
    year      = {2025},
    url       = {https://bmva-archive.org.uk/bmvc/2025/assets/papers/Paper_644/paper.pdf}
}

@article{seeground,
    title={Zero-Shot 3D Visual Grounding from Vision-Language Models},
    author={Rong Li and Shijie Li and Lingdong Kong and Xulei Yang and Junwei Liang},
    journal={ArXiv},
    year={2025},
    volume={abs/2505.22429},
    url={https://api.semanticscholar.org/CorpusID:278959500}
}

@inproceedings{scannet,
    title={ScanNet: Richly-annotated 3D Reconstructions of Indoor Scenes},
    author={Dai, Angela and Chang, Angel X. and Savva, Manolis and Halber, Maciej and Funkhouser, Thomas and Nie{\ss}ner, Matthias},
    booktitle = {Proc. Computer Vision and Pattern Recognition (CVPR), IEEE},
    year = {2017}
}

@InProceedings{vlsat,
    author    = {Wang, Ziqin and Cheng, Bowen and Zhao, Lichen and Xu, Dong and Tang, Yang and Sheng, Lu},
    title     = {VL-SAT: Visual-Linguistic Semantics Assisted Training for 3D Semantic Scene Graph Prediction in Point Cloud},
    booktitle = {Proceedings of the IEEE/CVF Conference on Computer Vision and Pattern Recognition (CVPR)},
    month     = {June},
    year      = {2023},
    pages     = {21560-21569}
}

@article{fuzzycateg1,
  title={Does Your 3D Encoder Really Work? When Pretrain-SFT from 2D VLMs Meets 3D VLMs},
  author={Li, Haoyuan and Zhou, Yanpeng and Gao, Yufei and Tang, Tao and Han, Jianhua and Yuan, Yujie and Chen, Dave Zhenyu and Bian, Jiawang and Xu, Hang and Liang, Xiaodan},
  journal={arXiv preprint arXiv:2506.05318},
  year={2025}
}

@inproceedings{fuzzycateg2,
  title={Escaping Plato's Cave: Towards the Alignment of 3D and Text Latent Spaces},
  author={Hadgi, Souhail and Moschella, Luca and Santilli, Andrea and Gomez, Diego and Huang, Qixing and Rodol{\`a}, Emanuele and Melzi, Simone and Ovsjanikov, Maks},
  booktitle={Proceedings of the Computer Vision and Pattern Recognition Conference},
  pages={19825--19835},
  year={2025}
}

@inproceedings{fuzzyrel,
  title={Inst3d-lmm: Instance-aware 3d scene understanding with multi-modal instruction tuning},
  author={Yu, Hanxun and Li, Wentong and Wang, Song and Chen, Junbo and Zhu, Jianke},
  booktitle={Proceedings of the Computer Vision and Pattern Recognition Conference},
  pages={14147--14157},
  year={2025}
}

@article{sg20,
  title={Language-Grounded Dynamic Scene Graphs for Interactive Object Search With Mobile Manipulation},
  author={Daniel Honerkamp and Martin Buchner and Fabien Despinoy and Tim Welschehold and Abhinav Valada},
  journal={IEEE Robotics and Automation Letters},
  year={2024},
  volume={9},
  pages={8298-8305},
  url={https://api.semanticscholar.org/CorpusID:268379149}
}

@INPROCEEDINGS{sg53, 
    AUTHOR    = {Abdelrhman Werby AND Chenguang Huang AND Martin Büchner AND Abhinav Valada AND Wolfram Burgard}, 
    TITLE     = {{Hierarchical Open-Vocabulary 3D Scene Graphs for Language-Grounded Robot Navigation}}, 
    BOOKTITLE = {Proceedings of Robotics: Science and Systems}, 
    YEAR      = {2024}, 
    ADDRESS   = {Delft, Netherlands}, 
    MONTH     = {July}, 
    DOI       = {10.15607/RSS.2024.XX.077} 
}

@inproceedings{sg17,
    title={Conceptgraphs: Open-vocabulary 3d scene graphs for perception and planning},
    author={Gu, Qiao and Kuwajerwala, Ali and Morin, Sacha and Jatavallabhula, Krishna Murthy and Sen, Bipasha and Agarwal, Aditya and Rivera, Corban and Paul, William and Ellis, Kirsty and Chellappa, Rama and others},
    booktitle={2024 IEEE International Conference on Robotics and Automation (ICRA)},
    pages={5021--5028},
    year={2024},
    organization={IEEE}
}

@article{sg34,
    title={Beyond Bare Queries: Open-Vocabulary Object Grounding with 3D Scene Graph},
    author={Sergey Linok and Tatiana Zemskova and Svetlana Ladanova and Roman Titkov and Dmitry A. Yudin and Maxim Monastyrny and Aleksei Valenkov},
    journal={2025 IEEE International Conference on Robotics and Automation (ICRA)},
    year={2024},
    pages={13582-13589},
    url={https://api.semanticscholar.org/CorpusID:272689035}
}

@inproceedings{sg16,
    title={Graphdreamer: Compositional 3d scene synthesis from scene graphs},
    author={Gao, Gege and Liu, Weiyang and Chen, Anpei and Geiger, Andreas and Sch{\"o}lkopf, Bernhard},
    booktitle={Proceedings of the IEEE/CVF Conference on Computer Vision and Pattern Recognition},
    pages={21295--21304},
    year={2024}
}

@article{sg59,
    title={Commonscenes: Generating commonsense 3d indoor scenes with scene graphs},
    author={Zhai, Guangyao and {\"O}rnek, Evin P{\i}nar and Wu, Shun-Cheng and Di, Yan and Tombari, Federico and Navab, Nassir and Busam, Benjamin},
    journal={Advances in Neural Information Processing Systems},
    volume={36},
    year={2024}
}

@article{ovsg,
    title={Context-aware entity grounding with open-vocabulary 3d scene graphs},
    author={Chang, Haonan and Boyalakuntla, Kowndinya and Lu, Shiyang and Cai, Siwei and Jing, Eric and Keskar, Shreesh and Geng, Shijie and Abbas, Adeeb and Zhou, Lifeng and Bekris, Kostas and others},
    journal={arXiv preprint arXiv:2309.15940},
    year={2023}
}

@inproceedings{3dgraphqa,
    title={3D Question Answering with Scene Graph Reasoning},
    author={Wu, Zizhao and Li, Haohan and Chen, Gongyi and Yu, Zhou and Gu, Xiaoling and Wang, Yigang},
    booktitle={ACM Multimedia 2024},
    year={2024},
}

@inproceedings{ffl3dog,
    title={Free-form description guided 3d visual graph network for object grounding in point cloud},
    author={Feng, Mingtao and Li, Zhen and Li, Qi and Zhang, Liang and Zhang, XiangDong and Zhu, Guangming and Zhang, Hui and Wang, Yaonan and Mian, Ajmal},
    booktitle={Proceedings of the IEEE/CVF international conference on computer vision},
    pages={3722--3731},
    year={2021}
}

@inproceedings{nyuv2,
    author    = {Nathan Silberman, Derek Hoiem, Pushmeet Kohli and Rob Fergus},
    title     = {Indoor Segmentation and Support Inference from RGBD Images},
    booktitle = {ECCV},
    year      = {2012}
}

@misc{quatrope,
      title={Scalable Object Relation Encoding for Better 3D Spatial Reasoning in Large Language Models}, 
      author={Shengli Zhou and Minghang Zheng and Feng Zheng and Yang Liu},
      year={2026},
      eprint={2603.24721},
      archivePrefix={arXiv},
      primaryClass={cs.CV},
      url={https://arxiv.org/abs/2603.24721}, 
}

@inproceedings{uni3d,
  title={Uni3d: Exploring unified 3d representation at scale},
  author={Zhou, Junsheng and Wang, Jinsheng and Ma, Baorui and Liu, Yu-Shen and Huang, Tiejun and Wang, Xinlong},
  booktitle={International Conference on Learning Representations (ICLR)},
  year={2024}
}

\newpage
\appendix

\section{Additional Experiments}

\subsection{Ablation Studies}

In this section, we provide additional ablation studies on the choice of loss and similarity function.

\textbf{The choice of loss function.} We choose weighted MSE loss as it is more robust when there is a large difference between the number of target and non-target objects (which is common for 3D-VL tasks). To validate its effectiveness, we also train CAPruner using binary cross-entropy (BCE) loss. The accuracy on the validation set when using the BCE loss is 19.72, which is much lower than the accuracy when trained with the weighted MSE loss (24.57), verifying the superiority of the weighted MSE loss.

\textbf{The choice of similarity function.} In this experiment, we compare the model's accuracy when inferred using category-based fuzzy matching or the cosine similarity between DINOv2 \cite{dinov2} / Uni3D \cite{uni3d} feature vectors and the Bert-Large-Uncased \cite{bert} embedding as the similarity function. The results in Tab. \ref{tab:app_feat_comp} demonstrate that when using the dot product of the  embedding vectors (i.e., cross-modal alignment scores) to measure similarity, the accuracy of the model deteriorates, indicating lower quality of the pruned scene graph. Such results further demonstrate the superiority of using category-based fuzzy matching as the similarity function.

\begin{table}[H]
  \centering
  \small
  \begin{tabular}{c|ccc}
    \toprule
    \textbf{Matching} & \textbf{ScanRefer} & \textbf{ScanQA} & \textbf{Multi3DRef} \\
    \textbf{Feature} & \textbf{Acc.@0.25} & \textbf{EM@1} & \textbf{F1@0.25} \\
    \midrule
    Category & 55.04 & 21.02 & 55.60 \\
    DINOv2 & 54.05 & 20.81 & 55.12 \\
    Uni3D & 54.53 & 20.79 & 55.25 \\ \bottomrule
  \end{tabular}
  \caption{Comparison on model's accuracy when inferred using category-based fuzzy matching or cosine similarity of DINOv2 \cite{dinov2} / Uni3D \cite{uni3d} feature vectors as the similarity function.}
  \label{tab:app_feat_comp}
\end{table}

\subsection{Generalizability Verification}

In this section, we perform an experiment to investigate cross-dataset generalization. We train CAPruner on a single dataset and directly evaluate it on all datasets without retraining. In Tab. \ref{tab:app_gen}, the row indicates the dataset used for training, the column indicates the dataset used for evaluation, and the scores are the accuracies of the model. From the results, we can observe that CAPruner trained on only one dataset can achieve performance very close to the model trained on all datasets across all evaluation settings. This clearly demonstrates that CAPruner has strong cross-dataset generalization ability, and the model trained on one dataset can transfer to other datasets effectively without retraining.

\begin{table}[!ht]
    \centering
    \small
    \begin{tabular}{c|ccc}
    \toprule
        \textbf{Training Set} & \textbf{ScanRefer} & \textbf{Multi3DRef} & \textbf{ScanQA} \\ \midrule
        ScanRefer & 21.80 & 27.53 & 22.48 \\ 
        Multi3DRef & 21.71 & 27.49 & 22.21 \\ 
        All & 21.85 & 27.61 & 22.74 \\ \bottomrule
    \end{tabular}
    \caption{The model's accuracy when trained using a single dataset or the combination of all datasets and tested on the validation sets.}
    \label{tab:app_gen}
\end{table}
\section{Qualitative Results}

\begin{table*}[tp]
    \hrule
    \vskip 0.10in
    \begin{minipage}{0.48\textwidth}
        \centering
        \textbf{Proximity-based KNN}
    \end{minipage}
    \begin{minipage}{0.48\textwidth}
        \centering
        \textbf{CAPruner (Ours)}
    \end{minipage}\hfill
    
    \vskip 0.10in
    \hrule
    \vskip 0.10in

    (a) [scene0011\_00] 3D VG: It is a gray trash can, the trash can sits in the corner by where the TV is.

    \begin{minipage}{0.47\textwidth}
        \centering
        \includegraphics[width=0.98\linewidth]{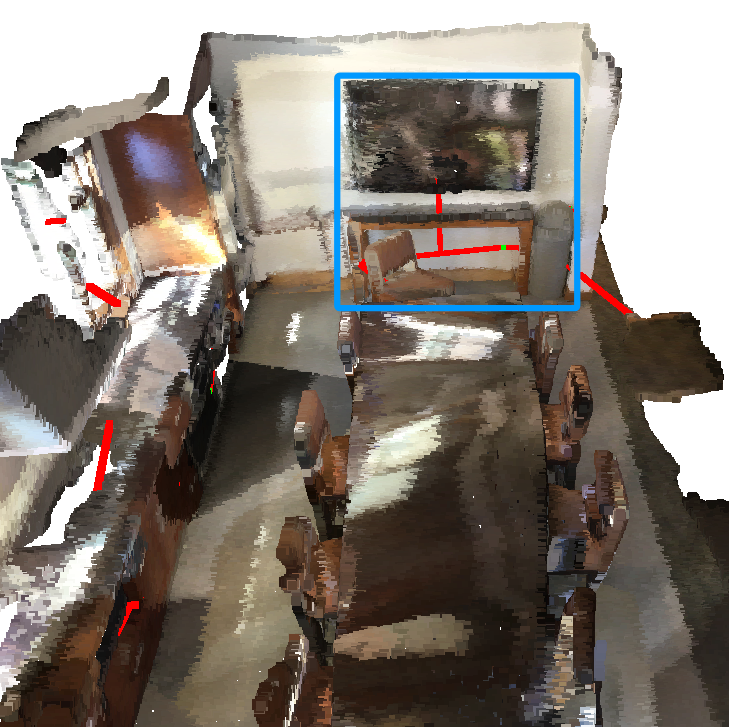}
    \end{minipage}
    \hfill
    \begin{minipage}{0.48\textwidth}
        \centering
        \includegraphics[width=0.98\linewidth]{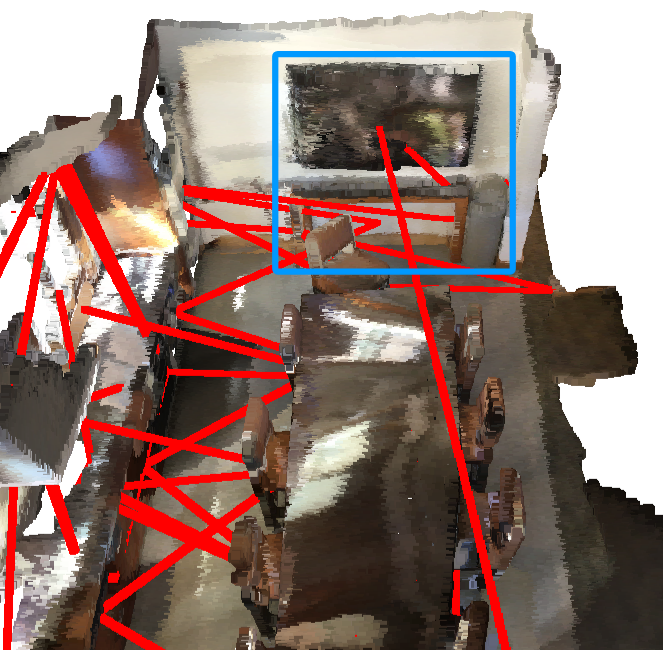}
    \end{minipage}
    
    \vskip 0.05in
    \hrule
    \vskip 0.10in

    (b) [scene0015\_00] 3D VG: It is a long brown table located opposite the crossed table on the other side.

    \begin{minipage}{0.45\textwidth}
        \centering
        \includegraphics[width=0.98\linewidth]{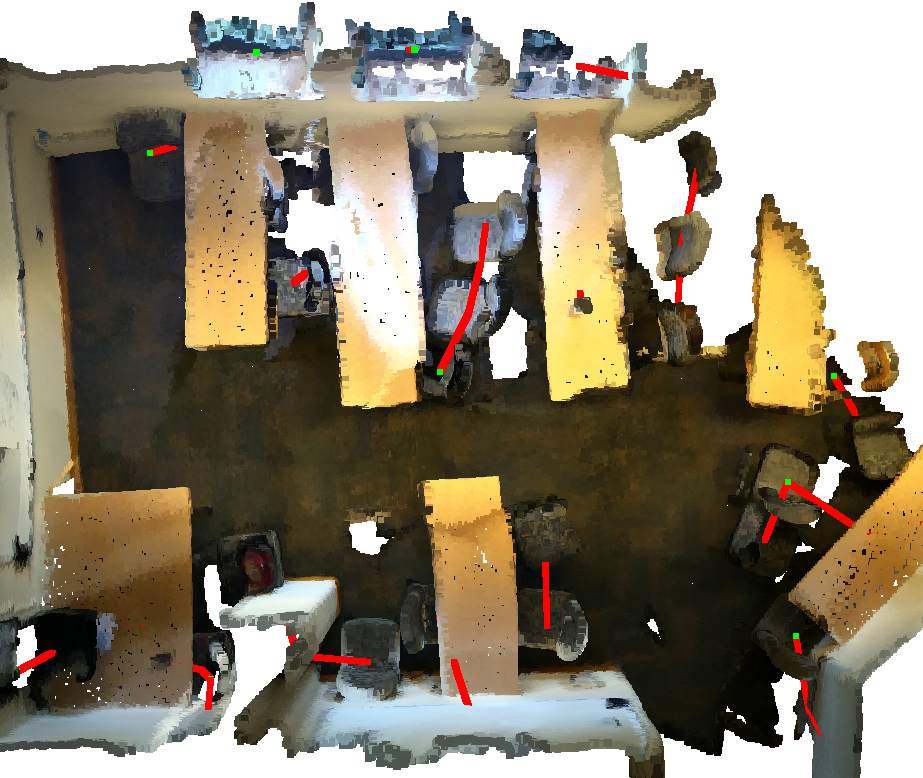}
    \end{minipage}
    \hfill
    \begin{minipage}{0.48\textwidth}
        \centering
        \includegraphics[width=0.98\linewidth]{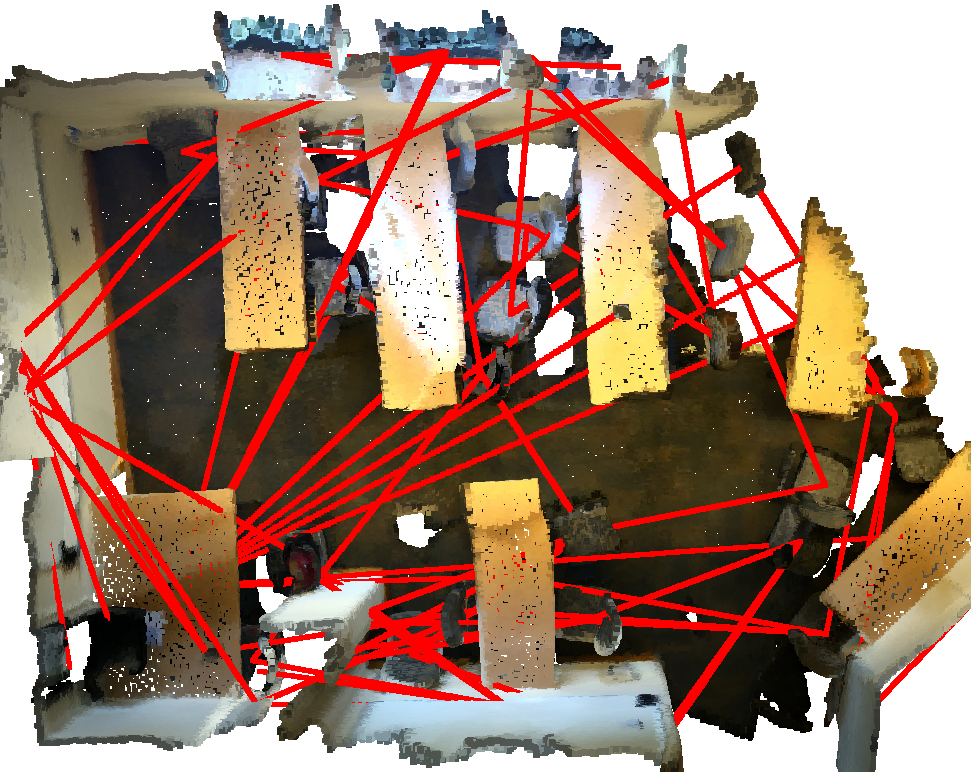}
    \end{minipage}
    
    \vskip 0.05in
    \hrule
    \vskip 0.10in

    (c) [scene0208\_00] 3D VG: It is a long brown table located opposite the crossed table on the other side.

    \begin{minipage}{0.48\textwidth}
        \centering
        \includegraphics[width=0.98\linewidth]{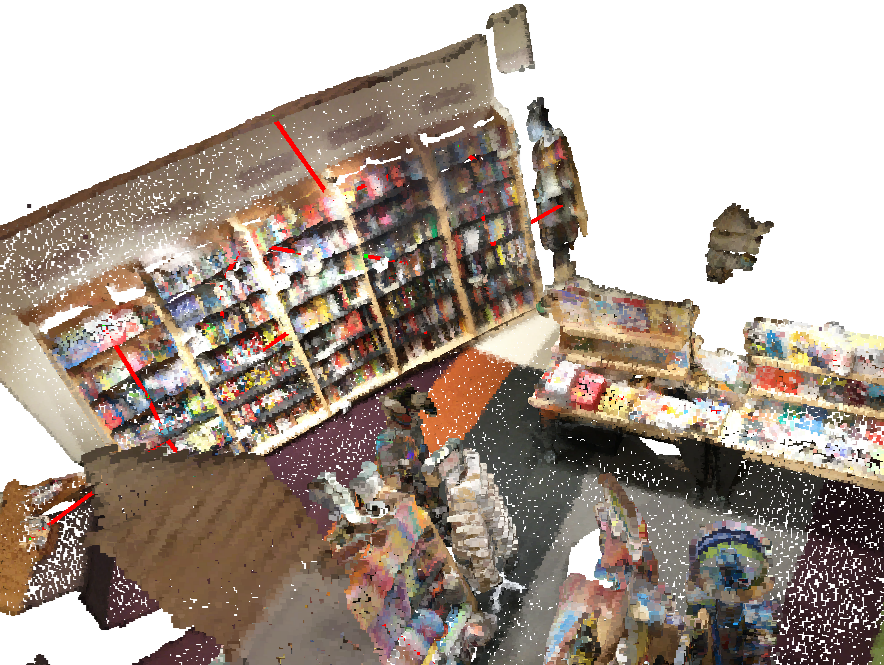}
    \end{minipage}
    \hfill
    \begin{minipage}{0.47\textwidth}
        \centering
        \includegraphics[width=0.98\linewidth]{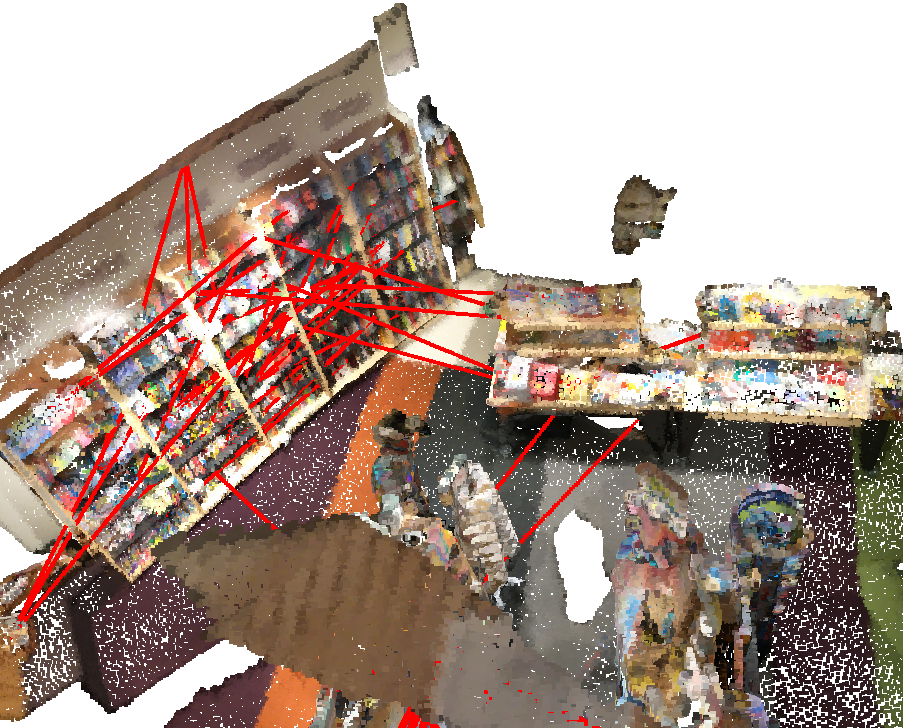}
    \end{minipage}
    
    \vskip 0.05in
    \hrule
    \vskip 0.10in
    
    \caption{Qualitative Results} \label{tab:app_qr}
\end{table*}

In this section, we visualize scene graphs pruned by CAPruner and compare them with scene graphs obtained through proximity-based KNN pruning. The context is chosen from the text descriptions in the 3D visual grounding (3D VG) dataset ScanRefer \cite{scanrefer}. The red edges in the figure correspond to those in the pruned scene graph. To enhance the clarity of visualization, we adopt a setting where only one incident edge with the highest weight (for CAPruner) / shortest distance (for proximity-based KNN) is retained.

\textbf{Case (a)} When using proximity-based KNN to prune the scene graph, the resulting graph on the left is overall sparse, hindering the model's ability to perceive the scene layout. In contrast, by introducing more edges with larger lengths, CAPruner on the right can better shape the structure of the scene. Moreover, when handling the task of locating ``it is a gray trash can, the trash can sits in the corner by where the TV is'', the KNN-pruned scene graph in the left figure fails to represent the spatial relation between the trash can and the TV, as the trash can are connected with the table and the wall, which are closer to it. By considering the categories of objects and the semantics of the task description, CAPruner gets rid of the limitations of spatial proximity and retains the edge between the trash can and the TV to reach the spatial reasoning requirement of the description.

\textbf{Case (b)} While proximity-based KNN pruning on the left can only focus on local spatial relations, CAPruner on the right can focus on long-range relations according to the requirement of the task. Such a characteristic enables the model to handle spatial relations with longer distances, e.g., ``opposite to'' and ``farthest''.

\textbf{Case (c)} As shown in the lower part of Fig. \ref{fig:findings} (b) and Case (c), there are lots of edges that have both ends within the same bookshelf. Such a phenomenon is caused by the preliminary step for LLM spatial reasoning, i.e., scene instance segmentation. When the segmentation model, e.g., Mask3D \cite{mask3d}, segments large objects (e.g., bookshelf) into multiple small objects (e.g., books or sections of the bookshelf). Such a segmentation makes proximity-based KNN prone to preserving edges between different parts of the same object. In contrast, because CAPruner also considers semantic similarity, the pruned scene graph is more robust to segmentation results.

\newpage

\section{Details for Datasets}

The number of examples in the datasets used for training and validation is as follows:

\begin{table}[!ht]
    \centering
    \begin{tabular}{ccc}
    \toprule
        Dataset & Training & Validation \\ \midrule
        ScanRefer & 32338 & 9508 \\ 
        ScanQA & 26138 & 4675 \\ 
        Multi3DRef & 37695 & 11120 \\ 
        SQA3D & 26623 & 3261 \\ \bottomrule
    \end{tabular}
    \caption{Number of examples in the datasets.}
    \label{tab:app_dataset}
\end{table}

\end{document}